\documentclass[runningheads]{llncs}
\usepackage[pdftex]{graphicx}
\usepackage{url}
\usepackage{comment}
\usepackage{color}
\usepackage{amsmath}
\usepackage{amsfonts}
\usepackage{subfigure} 
\usepackage{bm,amsmath}
\usepackage[ruled]{algorithm2e}
\usepackage{multirow}
\usepackage{color}
\usepackage{amsfonts}
\usepackage{ulem}
\begin{document}
\title{BayesGrad: Explaining Predictions of \\Graph Convolutional Networks}
%
%
\author{Hirotaka Akita \inst{1}\thanks{The work was done while the author was an intern at Preferred Networks,
Inc.}\and
Kosuke Nakago \inst{2} \and
Tomoki Komatsu \inst{2} \and
Yohei Sugawara \inst{2} \and
Shin-ichi Maeda \inst{2} \and
Yukino Baba \inst{3} \and
Hisashi Kashima \inst{1}
}
\authorrunning{H. Akita et al.}
%
\institute{Kyoto University\\
\email{h\_akita@ml.ist.i.kyoto-u.ac.jp\\
kashima@i.kyoto-u.ac.jp
}
\and
Preferred Networks, Inc.\\
\email{\{nakago, komatsu, suga, ichi\}@preferred.jp}
\and
{University of Tsukuba}\\
\email{baba@cs.tsukuba.ac.jp}
}
\maketitle
\begin{abstract}

Recent advances in graph convolutional networks have significantly improved the performance of chemical predictions, raising a new research question: ``how do we explain the predictions of graph convolutional networks?" A possible approach to answer this question is to visualize evidence substructures responsible for the predictions.
For chemical property prediction tasks, the sample size of the training data is often small and/or a label imbalance problem occurs, where a few samples belong to a single class and the majority of samples belong to the other classes. 
This can lead to uncertainty related to the learned parameters of the machine learning model.
To address this uncertainty, we propose BayesGrad, utilizing the Bayesian predictive distribution, to define the importance of each node in an input graph, which is computed efficiently using the dropout technique. We demonstrate that BayesGrad successfully visualizes the 
substructures responsible for the label prediction in the artificial experiment, even when the sample size is small. Furthermore, we use a real dataset to evaluate the effectiveness of the visualization. 
The basic idea of BayesGrad is not limited to graph-structured data and can be applied to other data types.

\end{abstract}

\section{Introduction}

The applications of deep neural networks are expanding rapidly in various fields, including chemistry and biology. 
Graph convolutional neural networks, which can handle graph-structured data (e.g., chemical compounds) as inputs, have opened the door to end-to-end learning for chemical prediction.
Many variants of graph convolutional neural networks have been proposed, which are now improving the performance of various chemical prediction tasks, including physical property prediction~\cite{mpnn}, toxicity prediction~\cite{weave}, solubility and drug efficiency prediction~\cite{nfp}, and total energy prediction~\cite{schnet}.

Deep neural networks automatically learn useful \textit{features} for prediction, which  sometimes outperform hand-engineered features carefully designed by domain experts,
enabling these neural networks to find new knowledge about molecular properties. However, the complex non-linear operations in deep neural networks make it prohibitively difficult to understand their behaviors.

{\it Sensitivity map}, also known as \textit{saliency map} or \textit{pixel attribution map}, is a common approach used to explain the reasons for the predictions of neural networks. 
The map assigns an \textit{importance score} to each substructure of an instance, which reflects the influence of the substructure on the final prediction, and visualizes high-scored substructures. 
The gradients are commonly used to measure the importance. 
A naive way entails using the size of the norm of the gradient~\cite{simonyanVZ13} (we call this approach \textit{VanillaGrad}). 
Sensitivity maps generated by this approach are typically noisy. As a result, SmoothGrad has been proposed to address this issue by adding noise to input samples and taking the mean values of the gradients~\cite{smoothgrad}.

The existing approaches do not take into account the uncertainty in the prediction of the model.
{The uncertainty becomes} particularly apparent in the chemical domain, because 
{the sample size of the chemical dataset is often small and/or a imbalance problem occurs, where only a few samples belong to a single class and the majority of the samples belong to the other classes.
{In such cases, it is difficult to estimate which substructures are responsible for a prediction.}
In this paper, we propose \textit{BayesGrad}, a novel sensitivity map algorithm that can deal with model uncertainty.
Our key idea is to quantify the uncertainty of a prediction utilizing its Bayesian predictive distribution. 
We implement the idea using the dropout, a common regularization technique for deep neural networks, because the outputs obtained using this technique approximate the expected value with respect to the Bayesian predictive distribution~\cite{Maeda14,dba}.
We conducted experiments using a synthetic compound dataset labeled with a particular substructure, and quantitatively evaluated the validity of our importance score. 
BayesGrad achieved superior performance, especially when the number of training data is small. 
We also use real datasets to visualize the bases of the predictions, and found that the visualized substructure is consistent with the known results.
Although we present the formalization of BayesGrad in the context of graph-structured data and demonstrate its efficiency in the chemical domain, BayesGrad is a general framework and, thus, can be applied to other data types such as images.

Our contributions to the literature are summarized as follows:

\begin{enumerate}
\item 
{\bf 
Bayesian approximation for sensitivity map visualization:} 
We propose a novel method that uses the dropout technique to quantify model uncertainty.
\item 
{\bf Application of gradient-based sensitivity map visualization for graphs:}
Most of the existing gradient-based sensitivity map algorithms are evaluated on image classification tasks. 
\item
{\bf Quantitative evaluation in the chemical domain:}
We quantitatively evaluated the performance of the gradient-based method to visualize the basis of a prediction.
\end{enumerate}

\section{Preliminaries}
We begin with the problem setting for sensitivity map visualization in graph prediction, followed by a brief review of several existing sensitivity map generation methods.
\subsection{Problem definition}
We assume that we have an (already trained) regression or classification model $f : {\mathcal G} \rightarrow \mathbb{R}$, where $G=\left(V,E\right) \in {\mathcal G}$ is a graph consisting of a set of nodes $V$ and a set of edges $E$, and the output of the model indicates the regression result or the classification score. Note that in the case of the binary classification model, the output of $f$ is the raw score in $\mathbb{R}$, not a value transformed by the sigmoid function. 
In the graph neural network $f$, a node $v_i$ is associated with a feature vector $\phi_i$. 

Given the model $f$ and a target input graph $G$, our goal is to assign an importance score $s_i$ to each node $v_i \in V$.  

\subsection{VanillaGrad}
 
There have been several recent attempts to interpret the predictions made by complex neural network models. Although these methods focus on images, they are easily applied to graphs. 
The gradient of $f$ with respect to feature $\phi_i$ is often used as the importance score of an input (i.e., a node) ~\cite{simonyanVZ13}:
%
\begin{align}
s_i = \left\|
\frac{\partial f(\phi; W)}{\partial\phi_i}
\right\|.
\label{eq:vanilla}
\end{align}
To simplify the calculation, we often use the 2-norm, but we can also use another norm such as the 1-norm.
We call the importance score defined in Eq. \eqref{eq:vanilla} \textit{VanillaGrad}.

\subsection{SmoothGrad}
It is known that sensitivity maps generated by VanillaGrad are likely to be noisy.
To address the problem, SmoothGrad~\cite{smoothgrad}  calculates the expected value of the gradient \eqref{eq:vanilla} over the Gaussian noise added to the input:
 \begin{align}
 \label{eq:valuewithnoise}
 s_i = E_{\epsilon \sim \mathcal{N}\left(0,\sigma^2\right)} \left[\left\|
 \frac{\partial f(\phi+\epsilon; W)}{\partial\phi_i}
 \right\| \right].
 \end{align}


We approximate the value of Eq. \eqref{eq:valuewithnoise} using sampling. SmoothGrad first generates $M$ noisy inputs $\{\check{\phi}_{i}^{m}\}_{m=1}^{M}$ by adding noise to the original input $\phi_i$; that is, $\check{\phi}_{i}^{m}$ is given as:
\begin{align}
\check{\phi}_{i}^{m} = \phi_{i} + \epsilon^{m},
\label{eq:smooth_dist}
\end{align}
where $\epsilon ^m$ is a sample from a Gaussian distribution $\mathcal{N}\left(0,\sigma^2\right)$
with a mean of zero a variance of $\sigma^2$. 
The importance score of a noisy input $\check{\phi}_{i}^{m}$ is then calculated as
\begin{align}
\displaystyle
\check{s}_i^m = \left\|
\frac{\partial f(\check{\phi}^m; W)}{\partial\phi_i}
\right\|.
\end{align}
Finally, the importance score of the original input $s_i$ is estimated as the  average of $\{\check{s}_i^m\}_{m=1}^M$:	
\begin{align}
s_i \approx \frac{1}{M}\sum_{m=1}^{M} \check{s}_i^{m},
\label{eq:smooth}
\end{align}
which is called \textit{SmoothGrad}.
Note that both of the variance of the Gaussian noise $\sigma^2$ and
the sample size $M$ are hyperparameters to be tuned.
In the original paper, $\sigma$ is tuned as a relative scale from the range of the input value $\phi$. However we used a fixed value of $\sigma$ for each input, because this was more stable in our experiment.

\subsection{Importance score calculation using signed values \label{sec_multiplying_gradient}}
In the previous discussion, the sensitivity map only gives how much each atom impacts on the prediction, but does not give whether the atoms have positive or negative effects on the prediction.
To address this, Shrikumar {\it et al.}~\cite{multiplying_gradient} 
used the product of the input and the gradient instead of the norm to evaluate how the atoms affect the output: 
\begin{align} 
(\phi_i - b_i)^{\top} \frac{\partial f(\phi; W)}{\partial\phi_i},
\end{align}
where $b$ denotes the baseline vector.
The above formula represents the effect on function $f$ when we change the $i$-th input from $\phi_i$ to $b_i$. It can be understood as (the negative of) the first-order term of the Taylor expansion of $f$ at $\phi$, which is evaluated at $b$.
Note that there is a freedom of choice of the baseline $b$; it is often set to  $0$, which corresponds to a black image in the image domain.
As we discuss in the experimental section, this technique gives us richer information in certain cases. 

\section{Proposed Method}

Existing approaches do not consider the uncertainty of the prediction by the model.
To address this issue, we propose BayesGrad, which 
quantifies the uncertainty of the sensitivity map
using Bayesian inference.
We first describe the formulation of BayesGrad, and then explain the practical implementation using the dropout technique.

\subsection{BayesGrad}


The existing methods are formulated in the framework of the maximum likelihood estimation of the neural network parameter $W$.
However, the learned $W$ is not necessarily stable and can vary with addition or deletion of a small portion of the training data sample.

In our formulation, we consider the uncertainty of the parameter $W$ by using the posterior of the neural network parameter $p(W|\mathcal{D})$ given the training data $\mathcal{D}$.
We consider the expected value of the importance score 
with respect to $p(W|\mathcal{D})$:
\begin{align}
\overline{s}_i = E_{W \sim p(W|\mathcal{D})} \left[ 
\left\|
\frac{\partial f(\phi; W)}{\partial\phi_i}
\right\|
\right].
\label{eq:Expected mean}
\end{align}

\noindent
We approximate this using sampling as
\begin{align}
\overline{s}_i &\approx \frac{1}{M}\sum_{j=1}^{M} s_i^{(j)}
\label{eq:mean},
\end{align}
where $s_i^{(j)}$ is the $j$-th importance score computed from the $j$-th sample $ W^{(j)} \sim p(W|\mathcal{D})$ ($j=1,\cdots,M$) as
\begin{align}
\displaystyle
s_i^{(j)} = \left\|
\frac{\partial f(\phi; W^{(j)})}{\partial\phi_i}
\right\|.
\end{align}






We call the importance score computed by Eq.\eqref{eq:mean} \textit{BayesGrad}. BayesGrad has a sample size $M$ as a hyperparameter. 

\subsection{Dropout as a Bayesian Approximation}
In order to implement BayesGrad, we need to take samples from the posterior distribution $p(W|\mathcal{D})$.
In general, the exact computation of the posterior is intractable, in which case we resort to approximation methods such as Markov chain Monte Carlo methods or variational Bayesian approximations. 
In particular, we utilize the dropout technique which can be interpreted as a variational Bayesian method because of its relatively small computational cost.
Dropout is originally introduced as a regularization technique to prevent overfitting~\cite{hinton_dropout,JMLR:v15:srivastava14a}, but recent studies show that dropout can be viewed as a kind of variational Bayesian inference~\cite{Maeda14,dba}.
We use the ``Dropout as a Bayesian Approximation (DBA)" technique to calculate the uncertainty of the model using dropout.
It approximates the posterior distribution $p(W|\mathcal{D})$
using variational distribution $q(W;\eta)$, where $\eta$ is a parameter to best approximate $p(W|\mathcal{D})$.
In DBA, $q(W;\eta)$ has a special form.
$W \sim q(W;\eta)$ is given as an Hadamard product of an adjustable constant matrix $\tilde W$ and the random mask matrix, and the stochasticity lies in each element of the random mask matrix that take the values either zero or one, typically with equal probability.

\subsection{Comparison between SmoothGrad and BayesGrad}
In contrast with SmoothGrad that takes the expectation of the gradient over possible fluctuations in the input variable,
BayesGrad smooths gradients over fluctuations in the model parameter $W$ that  follows the (approximate) posterior distribution $p(W|\mathcal{D})$. 
Validity of adding the Gaussian noise to the input depends on the task.
In the image domain, even if some noise is added to the original image, the noisy image still looks similar to the original one, and is still considered natural.
However, in the chemical domain, the input is the feature vector of each atom  which is originally a discrete object; hence, the noisy input does not correspond to a real atom anymore. The similar discussion also applies in other domain as well, e.g., word embedding in natural language processing.

Another benefit of BayesGrad is emphasized when the training data is few.
Model training tends to be unstable in such cases, and the model predictions tend to be stochastic. 
BayesGrad can treat this type of uncertainty by exploiting the Bayesian inference.

Note that the idea behind SmoothGrad taking the expectation in the input space and that of BayesGrad taking the expectation in the model space are not mutually exclusive, and we can combine both techniques to calculate a sensitivity map (as {\it BayesSmoothGrad}).
\section{Experiments}
We demonstrate the effectiveness of our approach in the chemical domain, where the sample size could be small and there is high demand for substructure visualization.
We first validate the methods using a synthetic dataset where the ground-truth substructures correlated with the target label are known.
In addition, we demonstrate the method using the real datasets and discuss its effectiveness.

We used Chainer Chemistry which is an open-source deep learning framework providing major graph convolutional network algorithms~\cite{chainer_chemistry}. 
We slightly modified the neural fingerprint method~\cite{nfp} and the gated-graph neural network (GGNN)~\cite{gatedgraph} in the library by including the dropout function to perform BayesGrad.
Our code used in this experiment is available at \url{https://github.com/pfnet-research/bayesgrad}.
Please refer the code for how to reproduce our result, including the hyperparameter configuration.

\subsection{Quantitative Evaluation on Tox21 Synthetic Data}
Tox21~\cite{huang2016tox21challenge} is a collection of chemical compounds including $11{,}757$ training, $295$ validation, and $645$ test data samples.
Each compound is associated with some of $12$ toxicity type labels; we used only the training and validation data in our experiment since the test dataset has no label information.
Since the original tox21 dataset does not have the information of what substructure actually contribute to their toxicity labels, We first used synthetic labels to quantitatively evaluate the different evidence visualization methods.
We assigned the label $1$ to compounds that contain pyridine~($\text{C}_5\text{H}_5\text{N}$) and $0$ to the remainder, which resulted in $760$ label-$1$ compounds and $10{,}997$ label-$0$ compounds in the training dataset.

We trained the GGNN with the dropout function to predict whether the input compound contains pyridine.
The GGNN has a gating architecture that enables the model to set the weights for important information.
After training the model, the ROC-AUC scores for both the training and the validation data were as high as 0.99, which suggests that the model was successfully trained. 

The validation dataset was used for testing. There are $28$ molecules that contain a pyridine substructure in the validation dataset.
We expect atoms that belong to pyridine rings to have a higher importance score than the others; hence, we selected the atoms in descending order of the importance scores after calculating the importance scores by each method.
We calculated the gradient of the output of the pre-final layer just before applying the sigmoid function that gives probability values, because the sigmoid function squashes the gradient and therefore the performance became worse if we took the gradient after the sigmoid function.

\begin{figure}[t!]
\begin{center}
  \includegraphics[clip, width=10cm,height=6cm,bb=20 10 532 320]{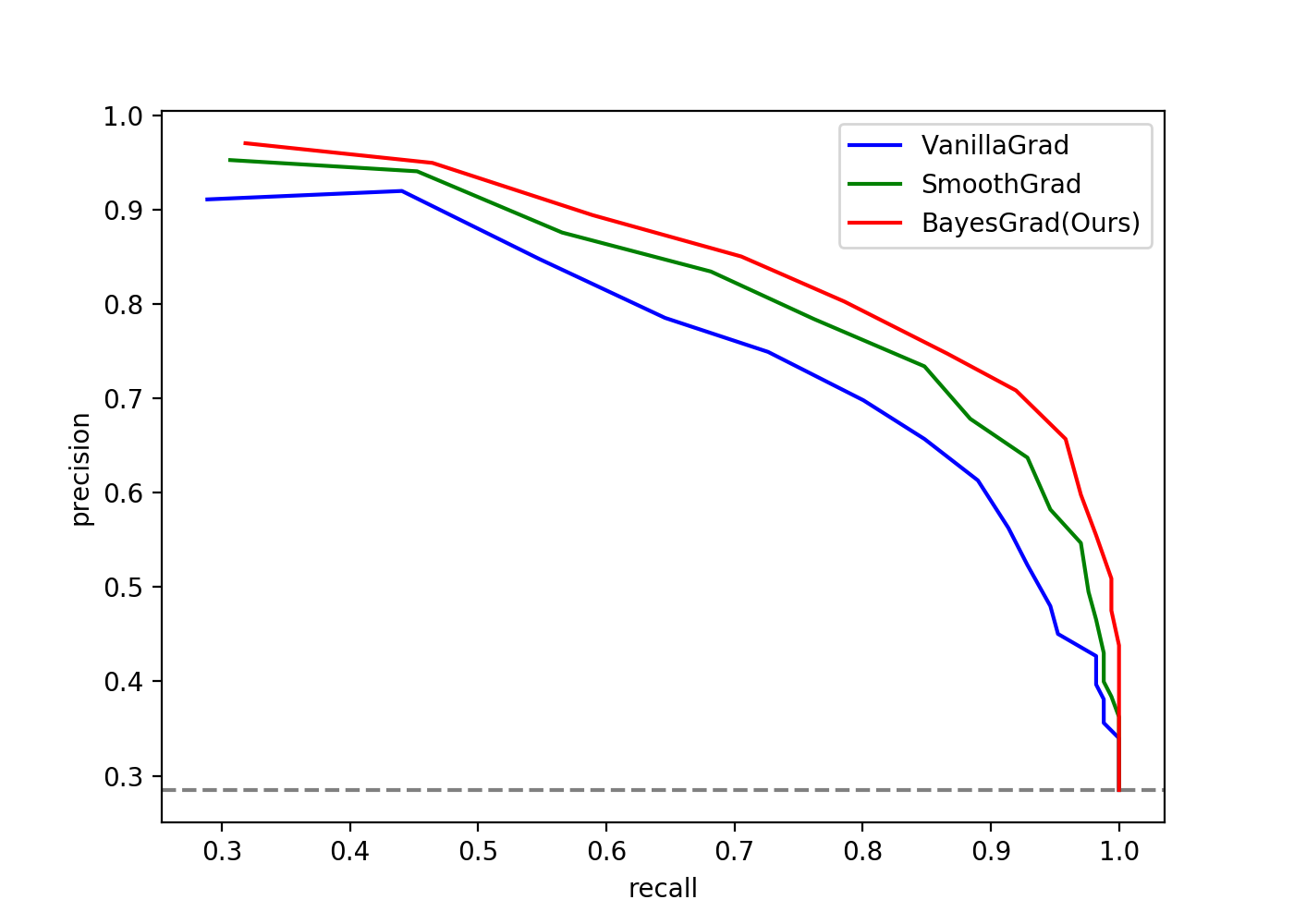}
  \caption{
  The precision-recall curve for each algorithm. All methods record high precision-recall curve.}
\label{fig:artificial}
\end{center}
\end{figure}
\begin{center}
\begin{figure}[t!]
    \subfigure[VanillaGrad]{%
        \includegraphics[width=4.1cm,height=2.5cm,bb=0 0 500 300]{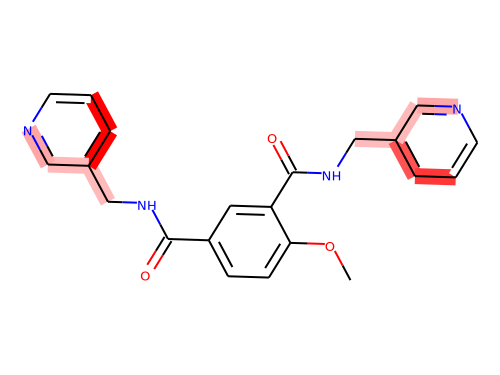}}%
    \subfigure[SmoothGrad]{%
        \includegraphics[width=4.1cm,height=2.5cm,bb=0 0 500 300]{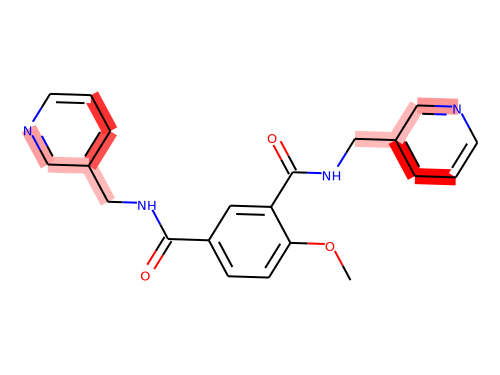}}%
    \subfigure[BayesGrad]{%
        \includegraphics[width=4.1cm,height=2.5cm,bb=0 0 500 300]{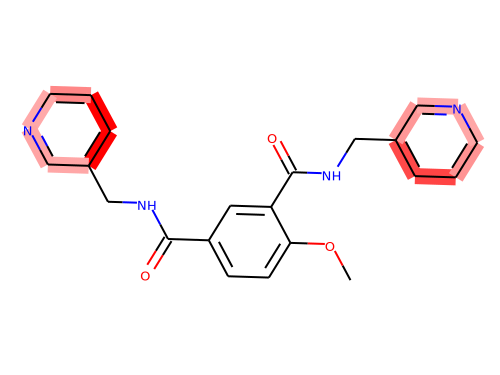}}%
    \caption{
    Examples of extracted substructure. The important atoms are highlighted. All methods successfully focuses on pyridine ($\text{C}_5\text{H}_5\text{N}$) substructure at the top-left and the top-right.
    }
\label{fig:example_pyridine}    
\end{figure}
\end{center}

Figure~\ref{fig:artificial} shows the precision-recall curve, 
where the precision indicates the proportion of the atoms consisting of pyridine rings in the extracted substructure, and the recall indicates the proportion of the extracted atoms in all the atoms in the pyridine rings. We used $M=100$ for the SmoothGrad and BayesGrad calculations.

\begin{table*}[htb]
\caption{PRC-AUC score between algorithms. The value before and after $\pm$ represent the mean and the standard deviation of PRC-AUC score calculated by 30 different models. Fixed value of $\sigma=0.15$ is used for SmoothGrad, $M=100$ is used for both SmoothGrad and BayesGrad.}
\label{table:prc-auc-pyridine}
\begin{center}
\begin{tabular}{lrrr}
\hline
Algorithm & PRC-AUC score \\
\hline
\hline
VanillaGrad & 0.506 $\pm$ 0.044 \\
\hline
SmoothGrad & 0.514 $\pm$ 0.042 \\
\hline
{\bf BayesGrad (Ours)} & {\bf 0.544 $\pm$ 0.019} \\
\hline
BayesSmoothGrad (Ours) & 0.536 $\pm$ 0.028 \\
\hline
\end{tabular}
\end{center}
\end{table*}

Figure~\ref{fig:example_pyridine} shows the sensitivity map visualization for each method.
All of the methods successfully extracted the substructure containing the pyridine ring.
This result implies that the gradient-based sensitivity map calculation is effective in extracting the substructure responsible for the target label in chemical prediction tasks.

Note that even though BayesGrad seems to outperform SmoothGrad or VanillaGrad in Fig.~\ref{fig:artificial}, this result is not deterministic owing to the stochastic behavior of SmoothGrad and BayesGrad.
To compare the performance of the methods,
 we consider a slightly difficult case with a small dataset.
This reflects a practical situation where limited data are available and the model's prediction tends to be uncertain.

To test that BayesGrad can deal with the uncertainty of the prediction,
we randomly select 30 different subset consisting of 1000 compounds from the original training dataset and obtained 30 different models.
We calculated the mean and standard deviation of their PRC-AUC scores.
The results are summarized in Table~\ref{table:prc-auc-pyridine}. BayesGrad records statistically higher PRC-AUC scores than both VanillaGrad and SmoothGrad.
We also tested BayesSmoothGrad method, which uses both dropout and noise; however, its performance did not improve in this experiment.

\subsection{Visualization on Tox21 Actual Data}
We also performed a toxicity prediction task experiment using the Tox21 dataset where each compound has some of 12 toxicity labels
We trained the prediction model for each of the labels, and visualized the grounds for prediction of the label \textit{SR-MMP} with the highest prediction accuracy ($0.889$ ROC-AUC in test data).

Figure~\ref{fig:tox21_srmmp} shows some interesting results;
Tyrphostin 9 (Fig.~\ref{fig:tox21_srmmp} (a)) is a tyrosine kinase inhibitor and is known to be a potent uncoupler of oxidative phosphorylation, which has a strong influence on the mitochondrial membrane potential~(\textit{SR-MMP}).
Terada \textit{et al.}~\cite{SF6847} examined the effect of the mitochondrial function of the acid-dissociable group using Tyrphostin 9 and a derivative, modified by methylation of its phenolic OH group. They confirmed that the acid-dissociable group is essential for uncoupling. We computed sensitivity maps for these compounds, as shown in Fig.~\ref{fig:tox21_srmmp} (a) and (b). Our visualization results are consistent with their experimental results. We also found similar compounds in the Tox21 dataset, with an acid-dissociable group, as shown in Fig.~\ref{fig:tox21_srmmp} (c) and (d). We confirmed that our visualization method has the potential to detect these essential substructures accurately.

\begin{figure}[t]
\begin{minipage}{0.5\hsize}
\begin{center}
    \subfigure[]{%
        \includegraphics[width=4.0cm,bb=0 0 500 400]{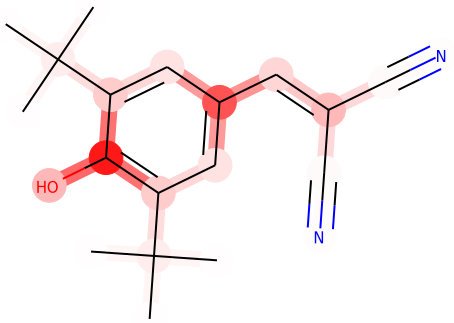}}%
\end{center}
\end{minipage}
\begin{minipage}{0.5\hsize}
\begin{center}
    \subfigure[]{%
        \includegraphics[width=4.0cm,bb=0 0 500 400]{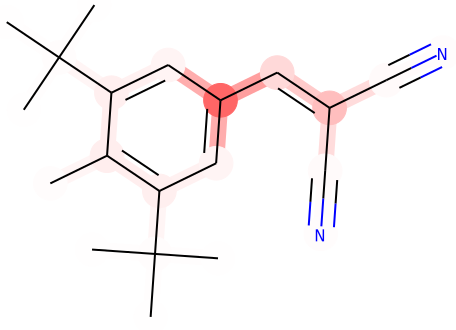}}%
\end{center}
\end{minipage}
\begin{minipage}{0.5\hsize}
\begin{center}
    \subfigure[]{%
        \includegraphics[width=4.0cm,bb=0 0 500 400]{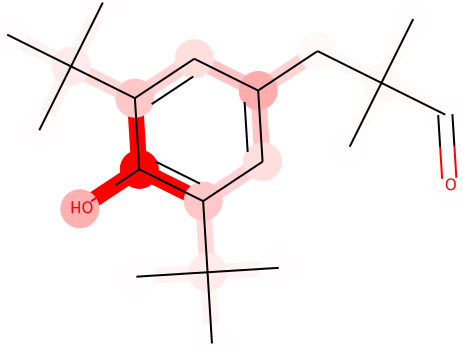}}%
\end{center}
\end{minipage}
\begin{minipage}{0.5\hsize}
\begin{center}
    \subfigure[]{%
        \includegraphics[width=4.0cm,bb=0 0 500 400]{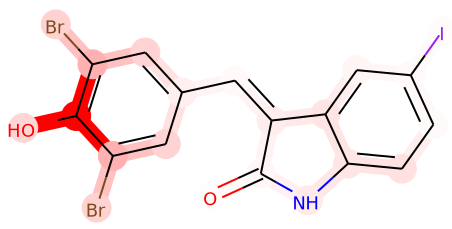}}%
\end{center}
\end{minipage}
\caption{
Visualizing the sensitivity map for SR-MMP toxicity prediction with BayesGrad.
(a) Chemical structure of Tyrphostin 9 and our model highlighted the phenolic OH (acid-dissociable) group, which is confirmed to be essential for uncoupling~\cite{SF6847}.
(b) O-methylated derivative of Tyrphostin 9 does not induce uncoupling. 
(c)(d) We found some compounds with similar sub-structure. Our model detected the same phenolic OH group.
}

\label{fig:tox21_srmmp}
\end{figure}

\subsection{Evaluation on Solubility Dataset}

\begin{figure}[t]
\begin{center}
  \includegraphics[clip, width=10cm,height=6cm,bb=0 5 500 305]{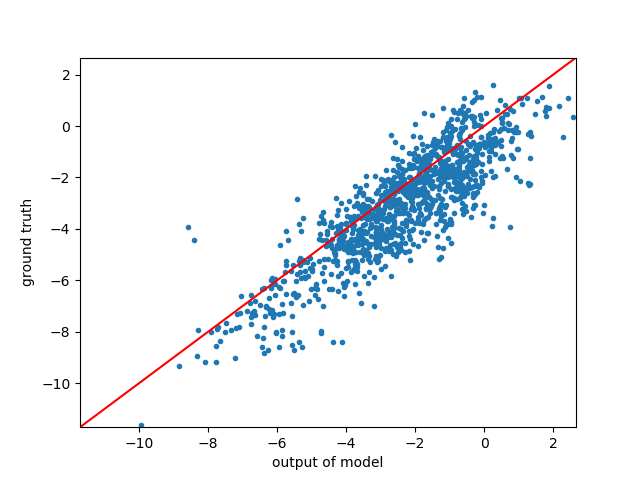}
  \caption{
  The scatter plot between measured solubility and our model output.
  The correlation coefficient is 0.851 and the mean absolute error is 1.024.}
\label{fig:solubility_result}
\end{center}
\end{figure}
Solubility is an important property in drug design because sufficient water-solubility is necessary for drug absorption. It is well known that some functional groups such as the hydroxyl group and primary amine group contribute to the hydrophilic nature, whereas other groups such as the phenyl group and ethyl group contribute to the hydrophobic nature. In addition, molecular weight has a strong correlation with solubility. Medicinal chemists need to modify the chemical structure by adding charged substituents, reducing the hydrophobic groups and the molecular weight, to improve solubility. However, if the molecule has a complicated structure, it becomes difficult to identify which part of structure is significant for the chemical property. Thus, our motivation is to provide a way to visualize which parts of a chemical structure are significant for the solubility. We demonstrated the effectiveness of our approach using a publicly available dataset.

We used the ESOL dataset~\cite{solubility_dataset} to evaluate our approach. This dataset contains 1{,}127 compounds with measured log solubility. 
We used the signed importance score explained in Section~\ref{sec_multiplying_gradient} to discriminate the positive/negative contributions to solubility.
The choice of the baseline $b$ is not trivial in chemical prediction tasks, where we consider the embedded feature vector space of atom representation as input. 
In our experiment, we used the baseline $b=0$, which corresponds to the mean of the prior distribution of the embedded feature vector. 
We used the neural fingerprint model~\cite{nfp} to evaluate this task. 

Figure~\ref{fig:solubility_result} shows the prediction result, where the  model achieved good performance for the solubility prediction.
Figure~\ref{fig:example_solubility} shows examples of the visualization for solubility prediction. 
Our approach accurately assigns positive importance scores to the hydrophilic atoms, and negative scores to the hydrophobic atoms, even for such compounds with  complicated structures.

\begin{center}
\begin{figure}[t!]
    \subfigure[]{%
        \includegraphics[width=4.1cm,height=2.5cm,bb=0 0 500 300]{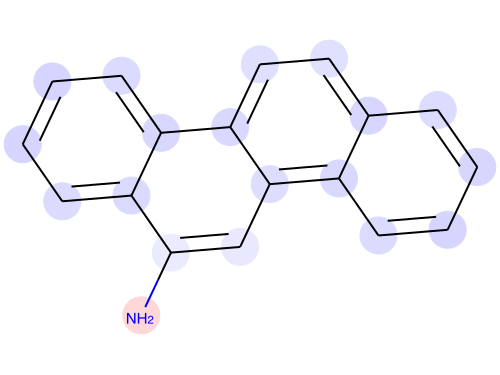}}%
    \subfigure[]{%
        \includegraphics[width=4.1cm,height=2.5cm,bb=0 0 500 300]{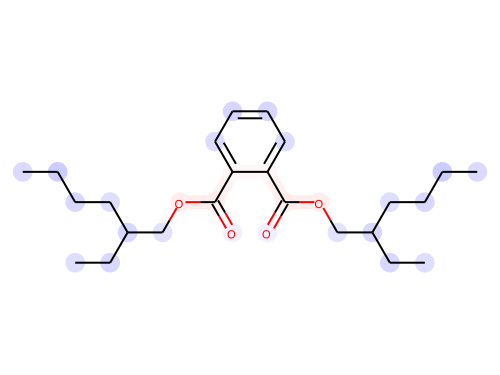}}%
    \subfigure[]{%
        \includegraphics[width=4.1cm,height=2.5cm,bb=0 0 500 300]{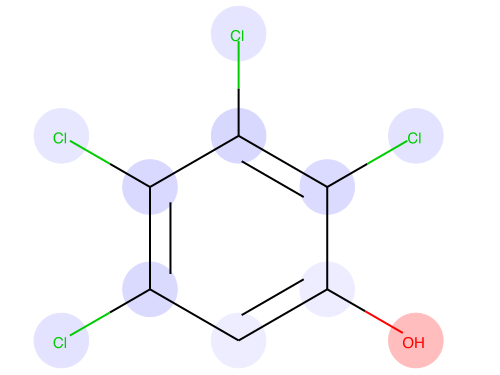}}%
    \caption{
Sensitivity maps for solubility prediction with BayesGrad. The atoms with  positive contributions to solubility are highlighted in red, and those with a negative contributions are highlighted in blue. 
Our results are mostly consistent with fundamental physicochemical knowledge.
(a) Positive score is assigned to a primary amine and negative scores is assigned to benzene rings, which are compatible with the facts that polycyclic aromatic hydrocarbon (PAH) has a hydrophobic nature and primary-amine is negatively charged, respectively.
(b) Ester is detected as the important substructure for hydrophilicity, which is known to have low polarity.
(c) Halogen substituents make a compound more lipophilic and less water-soluble. On the other hand, the hydroxyl group is negatively charged and expected to contribute to hydrophilicity.
    }
\label{fig:example_solubility}    
\end{figure}
\end{center}
\section{Conclusion}
We proposed a method to visualize a sensitivity map of chemical prediction tasks.
While existing methods focus on the visualization on image domain,
our quantitative evaluation with the tox21 dataset showed that BayesGrad outperforms the existing methods.
BayesGrad exploits the Bayesian inference technique to handle the uncertainty in predictions, which contributes to a robust sensitivity map, especially for small datasets.
Furthermore, we obtained the promising experimental results on the real datasets, which accord with the well-known chemical properties.

Elucidating the chemical mechanism is challenging research. We believe the proposed algorithm will lead to a better understanding of the chemical mechanism.
Our idea is easily applicable to other deep neural networks in other domains, which we leave to future research.

\bibliographystyle{unsrt}
\bibliography{reference}
\end{document}